\documentclass[10pt,twocolumn,letterpaper]{article}

\usepackage{cvpr}              

\usepackage{graphicx}
\usepackage{amsmath}
\usepackage{amssymb}
\usepackage{booktabs}
\usepackage{xcolor}
\usepackage{upgreek}
\usepackage[accsupp]{axessibility}  

\usepackage[pagebackref,breaklinks,colorlinks]{hyperref}

\usepackage[capitalize]{cleveref}
\crefname{section}{Sec.}{Secs.}
\Crefname{section}{Section}{Sections}
\Crefname{table}{Table}{Tables}
\crefname{table}{Tab.}{Tabs.}

\newcommand{\GITHUBLINK}{\href{http://sea-thru-nerf.github.io}{sea-thru-nerf.github.io}}

\def \bcB{{\beta^B}\xspace}
\def \bcD{{\beta^D}\xspace}

\def\cvprPaperID{31} 
\def\confName{CVPR}
\def\confYear{2023}

\newif\ifarxiv \newcommand{\arxiv}{\arxivtrue}
\newif\ifcamera 
\arxiv 

\begin{document}

\title{\emph{SeaThru}-NeRF: Neural Radiance Fields in Scattering Media\vspace{-0.3cm}}
\author{Deborah Levy$^{1}$, Amit Peleg$^{1}$, Naama Pearl$^{1}$, Dan Rosenbaum$^{1}$, Derya Akkaynak{$^{1,2}$},\\ Simon Korman$^{1}$, Tali Treibitz$^{1}$\\
$^{1}$University of Haifa , $^2$The Inter-University Institute for Marine Sciences in Eilat
\\
 {\texttt{\GITHUBLINK}}
 }



\twocolumn[{%
\renewcommand\twocolumn[1][]{#1}%
\maketitle

\begin{center}
    \centering
    \captionsetup{type=figure}
    \vspace{-0.32in}
    \includegraphics[width=1\textwidth]{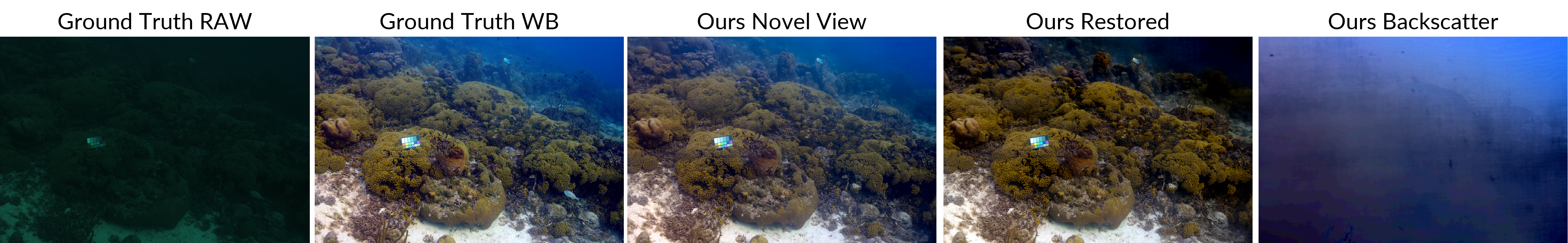} \vspace{-0.22in}
    \captionof{figure}{NeRFs have not yet tackled scenes in which the medium strongly influences the appearances of objects, as in the case of underwater imagery. By incorporating a scattering image formation model into the NeRF rendering equations, we are able to separate the scene into `clean' and backscatter components. Consequently, we can render photorealistic novel-views with or without the participating medium, in the latter case recovering colors as if the image was taken in clear air. 
    Results on the \textbf{Cura\c cao} scene: A RAW image (left) is brightened and white balanced (WB) for visualization, showing more detail, while areas further from the camera (top-right corner) are occluded and attenuated by severe backscatter - which is effectively removed in our restored image. Please zoom-in to observe the details.}\vspace{2pt}
    \label{fig:fig1}
\end{center}%
}]

\begin{abstract}\vspace{-0.1cm}

Research on neural radiance fields (NeRFs) for novel view generation is exploding with new models and extensions. However, a question that remains unanswered is what happens in underwater or foggy  scenes where the medium strongly influences the appearance of objects. Thus far, NeRF and its variants have ignored these cases. However, since the NeRF framework is based on volumetric rendering, it has inherent capability to account for the medium's effects, once modeled appropriately. We develop a new rendering model for NeRFs in scattering media, which is based on the SeaThru image formation model, and suggest a suitable architecture for learning both scene information and medium parameters.  We demonstrate the strength of our method using simulated and real-world scenes, correctly rendering novel photorealistic views underwater. Even more excitingly, we can render clear views of these scenes, removing the  medium between the camera and the scene and reconstructing the appearance and depth of far objects, which are severely occluded by the medium. Our code and unique datasets are available on the project's website.

\end{abstract}

\vspace{-0.4cm}
\section{Introduction}
\label{sec:intro}


The pioneering work of Mildenhall et al.~\cite{mildenhall2020nerf} on Neural Radiance Fields (NeRFs) has tremendously advanced the field of Neural Rendering,  due to its flexibility and unprecedented quality of synthesized images. 
Yet, the formulations of the original NeRF~\cite{mildenhall2020nerf} and its followup variants assume that images were acquired in clear air, i.e., in a medium that does not scatter or absorb light in a significant manner and that the acquired image is composed solely of the object radiance. 
The NeRF formulation is based on volumetric rendering equations that take into account sampled points along 3D rays. Assuming a clear air environment, an implicit assumption, which is often enforced explicitly with dedicated loss components \cite{barron2022mipnerf360}, is that a single opaque (high density) object is encountered per ray, with zero density between the camera and the object.

In stark contrast to clear air case, when the medium is \textit{absorbing} and / or \textit{scattering} (e.g., haze, fog, smog, and all aquatic habitats),
the volume rendering equation has a true volumetric meaning, as the entire volume, and not only the object, contributes to  image intensity. As the NeRF model estimates color and density at every point of a scene, it lends itself perfectly to general volumetric rendering, given that the appropriate rendering model is used. Here, we bridge this gap with \emph{SeaThru-NeRF}, a framework that incorporates a rendering model that takes into account scattering media.

This is achieved by assigning separate color and density parameters to the object (scene) and the medium, within the NeRF framework. Our approach adopts the SeaThru underwater image formation model~\cite{akkaynak2017space,akkaynak2018revised} to account for scattering media. SeaThru is a generalization of the standard wavelength-independent attenuation (e.g., fog) image formation model, where \textit{two} different wideband coefficients are used to represent the medium, which is more accurate when attenuation is wavelength-dependent (as in all water bodies and under some atmospheric conditions). 
In our model, the medium parameters are separate per color channel, and are learned functions of the  viewing angles, enforcing them to be constant only along 3D rays in the scene.

Attempting to optimize existing NeRFs on scenes with scattering medium results in cloud-like objects floating in space, while our formulation enables the network to learn the correct representation of the entire 3D volume, that consists of both the scene and the medium.
Our experiments demonstrate that \emph{SeaThru-NeRF} produces state-of-the-art photorealistic \textbf{novel view synthesis} on simulated and challenging real-world scenes (see Fig.~\ref{fig:fig1}) with scattering media, that include complex geometries and appearances. 
In addition, it enables:\\
\noindent 1. \textbf{Color restoration} of the scenes as if they were not imaged through a medium, as our modeling allows full separation of object appearance from the medium effects.\\
\noindent 2. \textbf{Estimation of 3D scene structure} which surpasses that of structure-from-motion (SFM) or current NeRFs, especially in far areas of bad visibility, as we jointly reconstruct and reason for the geometry and medium.\\ 
\noindent 3. \textbf{Estimation of wideband medium parameters}, which are informative properties of the captured environment, and potentially allowing simulation under different conditions.

\section{Related Work}
\label{sec:related}

\paragraph{Neural Radiance Fields (NeRFs):}

The original work on NeRFs~\cite{mildenhall2020nerf} has paved the way to a large capacity of followup work, with rapid and significant progress in many related aspects. For brevity, we focus in the following only on works closely related to ours, and refer the reader to a comprehensive review of the field~\cite{tewari2020state} prior to the introduction of NeRFs, and to~\cite{tewari2022advances} for the most recent.

NeRFs have been recently shown to be extremely powerful in multi-image settings that involve computational imaging tasks. These include 
HDR
~\cite{huang2021hdrnerf},
de-blurring
~\cite{ma2022deblur}, 
super-resolution
~\cite{wang2022nerf}, 
low-light enhancement
~\cite{mildenhall2022nerf} 
and denoising
~\cite{pearl2022nan}. The need to recover the clean `medium-free' version of images degraded by scattering and attenuation effects, likewise, can benefit from the neural rendering approach. 
SeaThru-NeRF is designed to \textit{model} the degradation, with the ability to recover its parameters,  and reconstruct the clean underlying scene and novel-view images.

Our work is related to recent efforts to improve the quality and robustness of NeRFs in challenging environments (e.g. ~\cite{guo2022nerfren,martin2021nerf, Niemeyer2020GIRAFFE}). 
Nerf-W~\cite{martin2021nerf} learns a per-image latent embedding that can capture appearance variations in complex scenes. It decomposes the scene into image-dependent and shared components to disentangle transient elements from the static scene.
%
%
NeRFReN~\cite{guo2022nerfren} is designed for scenes with reflections. It separates the scene as a sum of transmitted and reflected components, which are modeled as separate NeRFs.
Ref-NeRF~\cite{verbin2022ref} introduces a new parameterization and structuring of view-dependent appearance, which can represent  scenes with specularities and reflections.

Our approach has much in common with some of these methods, most notably the way of reconstructing the target image as a composition of components: `direct' and `backscatter' in our case, `transient' and `static' in~\cite{martin2021nerf}, or `transmitted' and  `reflected' in~\cite{guo2022nerfren}. 
However, as we demonstrate, these methods are limited on scenes in scattering media as they do not explicitly model the effects of the medium. Our method is different in that it can model a continuous medium component that is not an object.


\vspace{-0.4cm}
\paragraph{Graphics and Vision in Scattering Media:}\label{sec:related_UW}


Propagation of light in a medium is governed by the radiative transfer equation~\cite{chandrasekhar2013radiative}. This equation describes light interaction per particle in the medium and requires extensive Monte Carlo simulations for complete solutions~\cite{georgiev2013joint,jensen1998efficient,novak2018monte,pauly2000metropolis}, see~\cite{novak2018monte} for an excellent review. In many cases simplifying assumptions can be made about the medium that ease rendering~\cite{max2005local}, where the major one is single-scattering~\cite{pegoraro2009analytical,sun2005practical,walter2009single}. 
Realistic rendering requires having the scattering properties of the medium, which can be estimated in the lab~\cite{gkioulekas2013inverse,narasimhan2006acquiring}, or from \emph{in situ} images~\cite{bekerman2020unveiling}. An MLP for rendering synthetic atmospheric clouds 
was suggested in~\cite{kallweit2017deep}. 


In computer vision, image formation models under ambient illumination for bad weather~\cite{nayar1999vision} and underwater~\cite{schechner2005recovery} share the same general structure for the case of horizontal viewing. Artificial illumination used underwater~\cite{treibitz2009active} and in fog has additional terms that account for the nonuniformity of the source. Underwater, the medium parameters exhibit strong wavelength dependency that was shown to affect model accuracy in wideband camera channels. The SeaThru  model~\cite{akkaynak2017space,akkaynak2018revised,akkaynak2019sea} was suggested to overcome this issue.

Scene reconstruction in scattering media is an ill-posed problem which was initially solved with multiple frames~\cite{schechner2001instant,schechner2005recovery,narasimhan2002vision,treibitz2009active}. Later on, single-image methods have proposed a multitude of image priors to overcome the ill-posed nature of the problem, see~\cite{li2017haze,yang2019depth} for comprehensive reviews on single-image dehazing and underwater reconstruction, respectively. The introduction of deep-learning has led to an explosion of single-image dehazing and underwater reconstruction networks. See~\cite{sharma2021single} and~\cite{anwar2020diving} for recent reviews of deep-learning dehazing and underwater reconstruction works, respectively.

Underwater, it was suggested to solve for the 3D structure of the scene, prior to image restoration~\cite{bryson2016true,roser2014simultaneous} using the haze model and~\cite{akkaynak2019sea} using the revised one. In \emph{SeaThru-NeRF}, we simultaneously reconstruct the scene and its 3D structure, yielding multiple advantages that we demonstrate empirically. 
WaterNeRF~\cite{sethuraman2022waternerf} suggests an underwater neural renderer that estimates the medium parameters only with respect to histogram-equalized images, separately from the rendering. Here, we show the huge benefit of modeling the scattering medium within the rendering equations, over a variety of different scenes. 
Concurrently and independently to our work,~\cite{DehazeNeRF2023} presented a NeRF for haze. 

\section{Scientific Background}
\label{sec:background}

\subsection{Neural Radiance Fields (NeRFs)}
\label{sec:NERF}

The original NeRF formulation~\cite{mildenhall2020nerf} implicitly represents a 3D scene by a trainable continuous function. It is typically parameterized by an MLP 
    $f_\Theta: (\mathbf{x},\mathbf{d})\rightarrow(\mathbf{c},\sigma)$
which encodes the density $\sigma$ at the 3D point $\mathbf{x}=(x,y,z)$  and the color $\mathbf{c}=(r,g,b)$ emitted from this point in the viewing direction $\mathbf{d}=(\theta,\phi)$ (which is typically represented as a 2-element unit normalized 3D vector).\footnote{Vectors (3D coordinates or color components) are denoted in \textbf{bold}.}

This simple representation is used to simulate classical image-based rendering, by color accumulation along rays that are back-projected from a posed camera. If we parameterize points along a camera ray $\mathbf{r}$ by $\mathbf{r}(t)=\mathbf{o}+\mathbf{d}(t)$, where $\mathbf{o}$ is the camera center and $t\in\mathbb{R}_+$, the expected (image) color $C(\mathbf{r})$ along the ray can be written as:\footnote{Each color channel is integrated separately.}
\begin{equation}
C(\mathbf{r})=\int_{t_n}^{t_f}T(t)\sigma(t)\mathbf{c}(t)dt \label{eq.color_cont}
\end{equation}
where: 
integration along the ray is limited to the range $t\in[t_n,t_f]$ (near and far bounds on the scene contents); $\sigma(t)$ and $\mathbf{c}(t)$ are shorthands for the \textit{density} at the point $\mathbf{r}(t)$ and its emitted \textit{color} towards the camera center; $T(t)$ denotes the accumulated \textit{transmittance} along the ray from $t_n$ to $t$ (the probability that the ray travels from $t_n$ to $t$ without hitting other particles along the way), and is given by:
\begin{equation}
    T(t)=\exp\bigg(-\int_{t_n}^{t}\sigma(s)ds\bigg) \label{eq.transmittance}
\end{equation}

In practice, the rendered color $C(\mathbf{r})$ in Eq.~\eqref{eq.color_cont} is approximated with the quadrature rule, by discretizing the range $[t_n,t_f]$ into a set of $N$ intervals $I_i=[s_i,s_{i+1}]$ (where $t_n=s_0<\dots<s_N=t_f$), assuming that the density $\sigma$ and color $\mathbf{c}$ are constant along each interval (e.g. by querying the model once per interval at its center-point).

In the discretized version of Eq.~\eqref{eq.color_cont}:
\begin{equation}
    {\hat{C}(\mathbf{r})=\sum_{i=1}^N C_i(\mathbf{r})} \label{eq.discretized_color_contrib}
\end{equation}
the contribution $C_i(\mathbf{r})$ of the interval $I_i$ is given by:
\begin{equation}    C_i(\mathbf{r})=\int_{s_i}^{s_{i+1}}T(t)\sigma_i\mathbf{c}_i dt=T(s_i)\Big(1-e^{-\sigma_i\delta_i}\Big)\mathbf{c}_i \label{eq.color_interval}
\end{equation}
where $\sigma_i$ and $\mathbf{c_i}$ are the (constant) density and color along the $i$th interval, whose length is $\delta_i=s_{i+1}-s_i$, and the transmittance $T(s_i)$ at the beginning of the interval is:
\begin{equation}
   T(s_i)=\exp\bigg(-\sum_{j=0}^{i-1}\sigma_j\delta_j\bigg) \ \label{eq.transmittance_interval}
\end{equation}

This fully differentiable NeRF model (with the discretized rendering scheme) is trained with a simple reconstruction loss 
\begin{equation}\vspace{-2pt}
   L=\sum_{\mathbf{r}\in R} \lVert\hat{C}(\mathbf{r})-C(\mathbf{r})\rVert^2\ \label{eq.nerf_loss}
\end{equation}
comparing each rendered training image pixel $\hat{C}(\mathbf{r})$ to its ground truth color ${C}(\mathbf{r})$. The model can then be used to synthesize photorealistic novel view images.


\subsection{Image Formation in Scattering Media}\label{sec:uw_back}

\begin{figure}[t]
  \centering
  \includegraphics[width=0.8\linewidth]{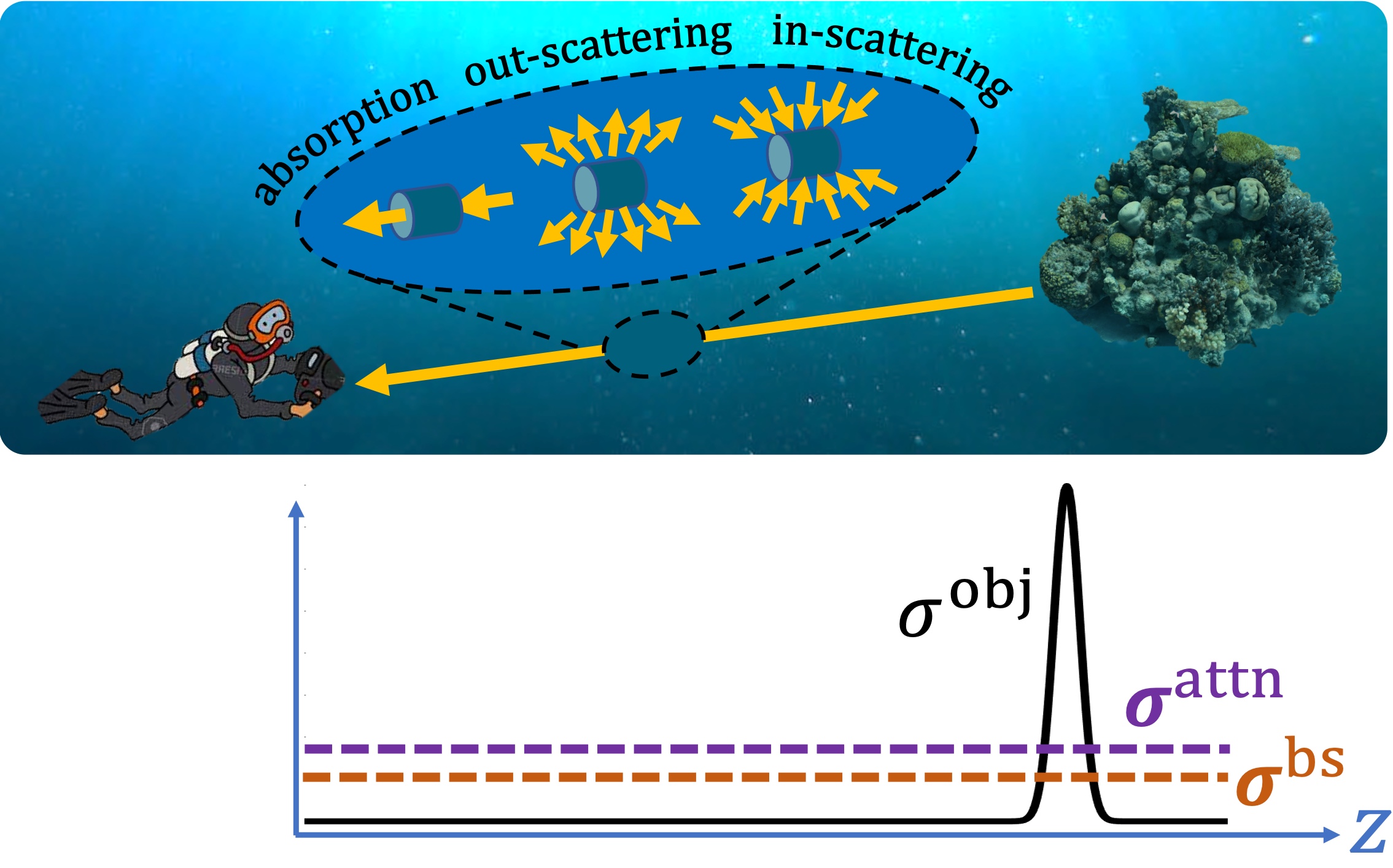} \vspace{-5pt}
   \caption{An illustration of our ray model. We assume a scene with at most a single opaque object at every ray. The medium is semi-transparent with constant density \emph{per-ray}. The densities governing the backscatter and the object in water are not the same~\cite{akkaynak2018revised}. }\vspace{-8pt}
   \label{fig:uwmodel}
\end{figure}

Image formation in fog, haze, or underwater differs from image formation in clear air in two major aspects. First, the \textit{direct} signal emanating from the object is attenuated as a function of distance and wavelength.
Second, this signal is occluded by \textit{backscatter} (termed also path-radiance or veiling-light) - radiance that is added due to the in-scattering from the particles along the line-of-sight (LOS), illustrated in Fig.~\ref{fig:uwmodel}. The intensity and color of the occluding backscatter layer are independent of the scene contents, and its intensity accumulates along the LOS, increasing with distance. As a result, the visibility and contrast of further objects is  significantly reduced and their colors are distorted.
 

We adopt the revised model~\cite{akkaynak2018revised} as the general model for image formation in scattering media under ambient illumination. Image intensity (per pixel, per color channel) is given as:

\vspace{-15pt}
\begin{equation}\vspace{-10pt}
I = 
\overbrace{\underbrace{J}_{\text{color}}\cdot
           \underbrace{(e^{-\bcD(\mathbf{v}_D)\cdot z})}_{\text{attenuation}}}^{\text{\textit{direct}}} 
           +
\overbrace{\underbrace{B^\infty}_{\text{color}}\cdot
           \underbrace{(1-e^{-\bcB(\mathbf{v}_B)\cdot z})}_{\text{attenuation}}}^{\text{\textit{backscatter}}}~         
           \label{eq.scattering}
           \vspace{10pt}
\end{equation}
where $I$ is the linear image captured by the camera of a scene with range $z$, $J$ is the clear scene that would have been captured had there been no medium along $z$, and $B^\infty$ is the backscatter water color at infinity, i.e., the backscatter at areas that contain no objects. Lastly, $\bcD$ and $\bcB$ are the attenuation and backscatter coefficients, respectively, the 
two parameters that describe the medium effects. 
The vectors $\mathbf{v}_D$ and \mbox{$\mathbf{v}_B$} represent the dependencies of $\bcD$ and $\bcB$ on range, object reflectance, spectrum of ambient light, spectral response of the camera, and the physical scattering and beam attenuation coefficients of the water body, all of which are functions of wavelength. 
It was shown in~\cite{akkaynak2019sea} that $\bcB$ can be assumed constant in an image, while $\bcD$ mainly depends on the object distance and weakly on object reflectance, thus solving for the full model requires at least 6 unknowns. The value of ${B^\infty}$ is usually assumed to be uniform in the scene but as discussed in~\cite{bekerman2020unveiling} it is almost never really uniform, because of the directionality of the sun, among other factors. 
 

This model is also applicable to haze and fog, where attenuation has very little dependence on wavelength. Then, image formation is greatly simplified because it can be assumed that there is only one medium parameter that is \emph{uniform} across the color channels, represented by a scalar ($\bcD=\bcB=\beta$)~\cite{schechner2001instant}. 
Underwater, often the simplifying (less accurate) assumption that $\bcD=\bcB$ is made, reducing to 3 unknowns~\cite{bekerman2020unveiling,berman2016non,schechner2005recovery}.

\section{\emph{SeaThru}-NeRF} 
\label{sec:our}

\subsection{Basic Model Derivation} \label{sec:basic_model}

Here we consider a more general setting than the original NeRF~\cite{mildenhall2020nerf}, in which light travels through a scattering \textit{medium} rather than free-space, resulting in a significant impact on the captured color. Following~\cite{max2005local} we suggest adding the medium to Eqs.~\eqref{eq.color_cont} and \eqref{eq.transmittance}:
\begin{equation}    C(\mathbf{r})=\int_{t_n}^{t_f}T(t)\Big(\sigma^{\rm obj}(t)\mathbf{c}^{\rm obj}(t)+\sigma^{\rm med}(t)\mathbf{c}^{\rm med}(t)\Big)dt \label{eq.color_cont_UW}
\end{equation}
where:
\begin{equation}
    T(t)=\exp\left(-\int_{t_n}^{t}\Big(\sigma^{\rm obj}(s)+\sigma^{\rm med}(s)\Big)ds\right)\;\;, \label{eq.transmittance_UW}
\end{equation}
using separate color and density parameters for the \textit{object} and \textit{medium}. 
Notice that these equations reduce to Eqs.~\eqref{eq.color_cont} and~\eqref{eq.transmittance} by simply considering the case where the medium density $\sigma^{\rm med}$ is zero.

Moving to the discretized version, we similarly get\footnote{Full derivations of Eqs.~(\ref{eq.color_interval_UW},~\ref{eq.transmittance_interval_UW}) is provided in the appendix.} the following generalizations of Eqs.~\eqref{eq.color_interval} and~\eqref{eq.transmittance_interval}:
\begin{equation}
C_i(\mathbf{r})=T(s_i)\left(1-e^{-(\sigma^{\rm obj}_i+\sigma^{\rm med}_i)\delta_i}\right)
\small
\frac{\sigma^{\rm obj}_i\mathbf{c}^{\rm obj}_i+\sigma^{\rm med}_i\mathbf{c}^{\rm med}_i}{\sigma^{\rm obj}_i+\sigma^{\rm med}_i}
\label{eq.color_interval_UW}
\end{equation}
with transmittance
\begin{equation}
   T(s_i)=\exp\bigg(-\sum_{j=0}^{i-1}(\sigma^{\rm obj}_j+\sigma^{\rm med}_j)\delta_j\bigg) \;\;. \label{eq.transmittance_interval_UW}
\end{equation}


Splitting the discretized color rendering equations (\ref{eq.discretized_color_contrib} ,\ref{eq.color_interval_UW} and \ref{eq.transmittance_interval_UW}) into the `object' and `medium' components, we get:
\begin{equation}
    {\hat{C}(\mathbf{r})=\sum_{i=1}^N \hat{C}^{\rm obj}_i(\mathbf{r})+\sum_{i=1}^N \hat{C}^{\rm med}_i(\mathbf{r})} 
    ~~~~~~~~~~~
    \label{eq.split_discretized_color_contrib}
\end{equation}
\begin{equation} 
\hat{C}^{\rm obj}_i(\mathbf{r})=T(s_i)
\left(1-e^{-(\sigma^{\rm obj}_i+\sigma^{\rm med}_i)\delta_i}\right)
\small\frac{\sigma^{\rm obj}_i\mathbf{c}^{\rm obj}_i}{\sigma^{\rm obj}_i+\sigma^{\rm med}_i}
\label{eq.color_interval_UW_obj}
\end{equation}
\begin{equation}
\hat{C}^{\rm med}_i(\mathbf{r})=T(s_i)
\left(1-e^{-(\sigma^{\rm obj}_i+\sigma^{\rm med}_i)\delta_i}\right)
\small\frac{\sigma^{\rm med}_i\mathbf{c}^{\rm med}_i}{\sigma^{\rm obj}_i+\sigma^{\rm med}_i}
\label{eq.color_interval_UW_med}
\end{equation}
As a first step towards constraining and simplifying our model, we constrain the medium parameters to be constant along 3D viewing rays. We drop the respective interval indices and remain with $\boldsymbol{\upsigma}^{\rm med}$ and ${\mathbf{c}^{\rm med}}$ that depend only on the ray $\mathbf{r}$. The ${\mathbf{c}^{\rm med}}$ uniformity
stems from the common assumption of a uniform phase function (the dependence of scattered radiance on scattering angle) along the LOS~\cite{novak2018monte,max2005local}.
Regarding the density $\boldsymbol{\upsigma}^{\rm med}$, we take it to be separate per color channel, but constant per ray. This is far less restrictive compared to models that assume constancy per image or even per scene~\cite{akkaynak2019sea}. These constraints will be enforced by respective structural choices in the network.

Furthermore, we assume that objects in the scene are  opaque, hence the object density along the ray is close to zero except for a high peak in the object location. On the other hand, the medium is semi-transparent, characterized by a low non-zero density. 
This implies that \mbox{$\boldsymbol{\upsigma}^{\rm med} \gg \sigma^{\rm obj}$} before the object and \mbox{$\boldsymbol{\upsigma}^{\rm med} \ll \sigma^{\rm obj}$} at the object, as illustrated in Fig.~\ref{fig:uwmodel}. 
Therefore, Eqs.~(\ref{eq.color_interval_UW_obj}-\ref{eq.color_interval_UW_med}) reduce to 
%
\begin{eqnarray}
  \hspace{-17pt} \hat{C}^{\rm obj}_i(\mathbf{r}) \hspace{-5pt}&=& \hspace{-5pt}
  T_i \cdot \Big(1-e^{-\sigma^{\rm obj}_i\delta_i}\Big) \cdot \mathbf{c}^{\rm obj}_i
\\
  \hspace{-17pt} \hat{C}^{\rm med}_i(\mathbf{r}) \hspace{-5pt}&=& \hspace{-5pt} 
  T_i\cdot \Big(1-e^{-\boldsymbol{\upsigma}^{\rm med}\delta_i}\Big) \cdot  \mathbf{c}^{\rm med}
\\
     \hspace{-17pt} T_i \hspace{-5pt}&=& \hspace{-5pt} \exp\Big(-\sum_{j=0}^{i-1}\sigma^{\rm obj}_j\delta_j\Big)\cdot \exp\left(-\boldsymbol{\upsigma}^{\rm med} s_i\right)
    \label{eqn.def_transmission}
\end{eqnarray}

\subsection{Relation to Underwater Image Formation} \label{sec:relation_to_UW_model}

In this section we show that our model can be reduced to the image formation model described in Sec.~\ref{sec:uw_back}, which is not based on volumetric rendering.
To this end we consider sampling the ray along intervals with constant size $\delta$, and the appearance of the  opaque object at a depth $z$, at the beginning of interval \mbox{$I_k=[s_k,s_{k+1}]$}, for some integer $k$ (that is: $s_i=i\cdot\delta$ for every $i$ and $z=k\cdot\delta$ in particular). 

In this case, $\sigma^{\rm obj}_i\approx 0$ for all $i<k$, and $\sigma^{\rm obj}_k\gg\boldsymbol{\upsigma}^{\rm med}$. This implies that $C^{\rm med}_k(\mathbf{r}) \ll C^{\rm obj}_k(\mathbf{r})$ and $C^{\rm obj}_i(\mathbf{r})\approx 0$, therefore we can write the rendering Eq. \eqref{eq.split_discretized_color_contrib} as \vspace{-0.2cm}
\begin{equation}
\hat{C}(\mathbf{r}) \approx 
\hat{C}^{\rm obj}_k(\mathbf{r}) + \sum_{i=0}^{k-1}\hat{C}^{\rm med}_i(\mathbf{r})\label{eq.split_discretized_rendering}\;\;.
\end{equation}
Furthermore, since the object is opaque, we can assume that the density $\sigma^{\rm obj}_k$ is large enough such that the transmittance at the end of the $k$th interval $T(s_{k+1})$ drops practically to zero, that is: $e^{-\sigma^{\rm obj}_k\delta}\approx 0$. Therefore
\begin{eqnarray}
\hat{C}^{\rm obj}_k(\mathbf{r}) \hspace{-7pt}&=& \hspace{-7pt}T_i \cdot \big(1-e^{-\sigma^{\rm obj}_i\delta}\big) \cdot \mathbf{c}^{\rm obj}_i\label{eq.c_k_obj}
\\
\hspace{-7pt}&\approx&\hspace{-7pt}  e^{-k\boldsymbol{\upsigma}^{\rm med}\delta}\big(1-e^{-\sigma^{\rm obj}_k\delta}\big)\mathbf{c}^{\rm obj}_k  \approx e^{-\boldsymbol{\upsigma}^{\rm med}\delta\cdot k}\mathbf{c}^{\rm obj}_k\nonumber
\end{eqnarray}
and 
\begin{eqnarray}
 \sum_{i=0}^{k-1}\hat{C}^{\rm med}_i(\mathbf{r}) \hspace{-5pt} &=& \hspace{-5pt}
\sum_{i=0}^{k-1} T_i\cdot \big(1-e^{-\boldsymbol{\upsigma}^{\rm med}\delta}\big) \cdot \mathbf{c}^{\rm med}
\label{eq.c_i_med}
 \\ 
\hspace{-5pt} &\approx& \hspace{-5pt} \sum_{i=0}^{k-1} e^{-i\boldsymbol{\upsigma}^{\rm med}\delta}\cdot\big(1-e^{-\boldsymbol{\upsigma}^{\rm med}\delta}\big)\cdot \mathbf{c}^{\rm med}~~~~\nonumber\\ 
\hspace{-5pt} &=& \hspace{-5pt} \big(1-e^{-\boldsymbol{\upsigma}^{\rm med}\delta\cdot k}\big)\cdot \mathbf{c}^{\rm med}~~~~\nonumber
\end{eqnarray}

Substituting Eqs.~\eqref{eq.c_k_obj} and \eqref{eq.c_i_med} into Eq.~\eqref{eq.split_discretized_rendering} (recalling that the depth $z$ equals $k\cdot\delta$) yields: 
\begin{equation}
      \hat{C}(\mathbf{r}) \approx \\     
                  {e^{-\boldsymbol{\upsigma}^{\rm med}z}}\cdot
                  {\mathbf{c}^{\rm obj}_k} +
                  \Big({1-e^{-\boldsymbol{\upsigma}^{\rm med}z}}\Big)\cdot
                  {\mathbf{c}^{\rm med}}  \;\;,    
\end{equation}
resulting precisely in the commonly used scattering media image formation model~\cite{berman2020underwater,schechner2005recovery}. This can be seen by comparing to Eq.~\eqref{eq.scattering} where $\boldsymbol{\upsigma}^{\rm med}=\bcD=\bcB$ plays the role of the attenuation coefficient (equal for direct and backscatter), $\mathbf{c}^{\rm med}$ is the veiling light $B^\infty$ (backscatter color at infinity) and ${\mathbf{c}^{\rm obj}_k}$ the clear image color $J$. 

\begin{figure}[t]
  \centering
  \includegraphics[width=0.95\linewidth]{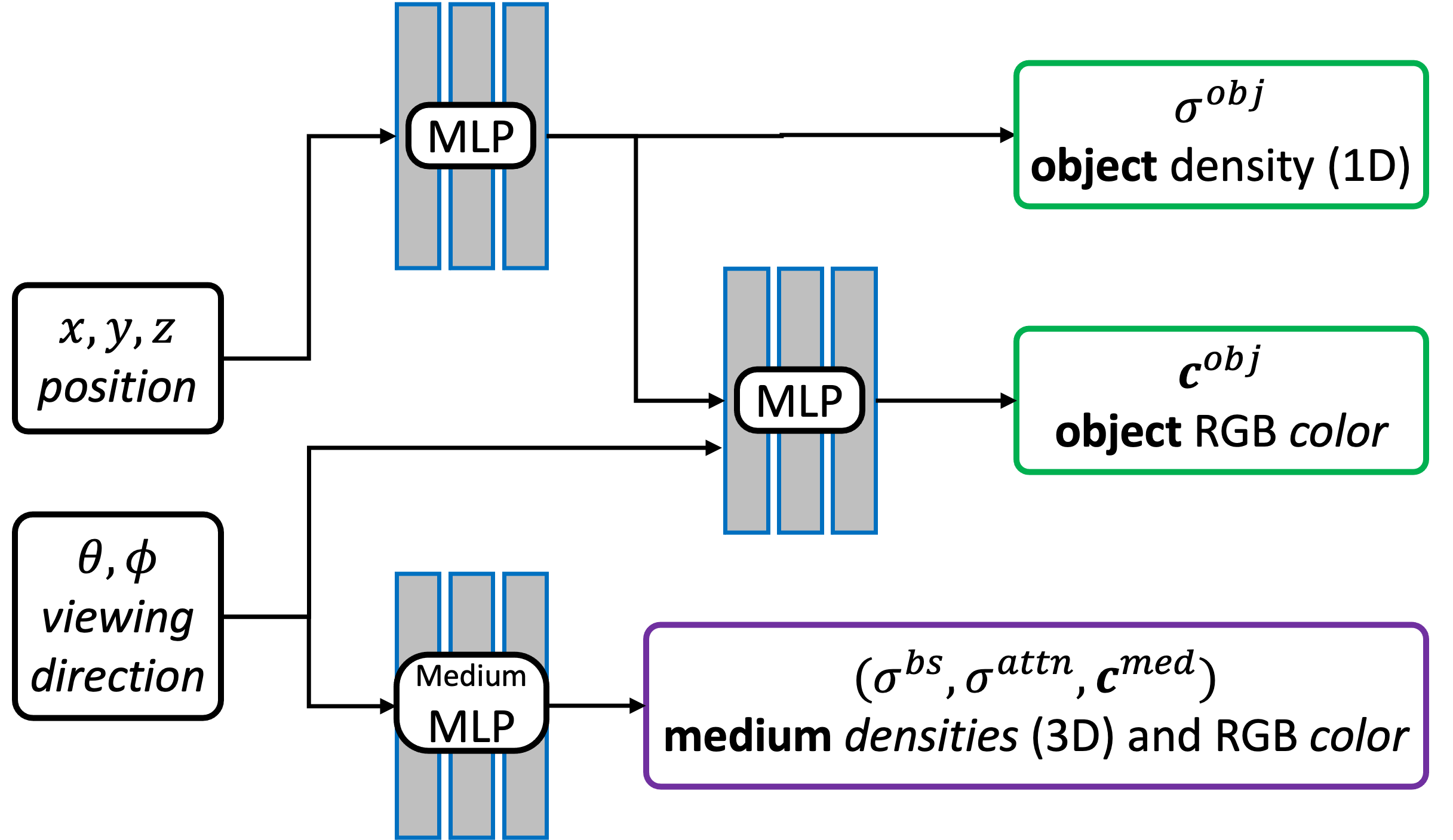} \vspace{-5pt}
   \caption{\textbf{SeaThru-NeRF architecture.} The computation of the `object' outputs (density and color, in green), follows the standard NeRF architecture, while the `medium' components (in purple) are computed once per ray by a separate subnet (the `medium MLP')  that depends only on the viewing direction.}
   \label{fig:arch}\vspace{-0.3cm}
\end{figure}

\subsection{Final Model} \label{sec.final_model}
We make several refinements with respect to the basic model presented in Section~\ref{sec:basic_model}.
In Sec.~\ref{sec:relation_to_UW_model}, we showed that our rendering equations lead to an image formation model with identical attenuation coefficients for the direct and backscatter components. Following the discussion in Sec.~\ref{sec:uw_back}, 
and as has been shown in~\cite{akkaynak2018revised}, when using such equations with a camera with wideband color channels, the 
effective $\boldsymbol{\upsigma}^{\rm med}$ that is experienced by the camera in $C^{\rm obj}_i(\mathbf{r})$ is different than the one experienced in $C^{\rm med}_i(\mathbf{r})$.
Therefore, in our final model we use different parameters for $\boldsymbol{\upsigma}^{\rm med}$ in each component and term them $\boldsymbol{\upsigma}^{\rm attn}$ and $\boldsymbol{\upsigma}^{\rm bs}$ for $C^{\rm obj}_i(\mathbf{r})$ and $C^{\rm med}_i(\mathbf{r})$ respectively. Our final equations are:
\begin{equation*}
  ~~\hat{C}^{\rm obj}_i(\mathbf{r}) =
  T^{\rm obj}_i\cdot \exp\left(-{\boldsymbol{\upsigma}^{\rm attn}} s_i\right)
  \cdot \big(1-\exp({-\sigma^{\rm obj}_i\delta_i})\big) \cdot \mathbf{c}^{\rm obj}_i~~~~~~
\end{equation*}
\begin{equation*}
  ~~\hat{C}^{\rm med}_i(\mathbf{r}) = 
  T^{\rm obj}_i\cdot \exp\left(-{\boldsymbol{\upsigma}^{\rm bs}} s_i\right)
  \cdot \big(1-\exp({-\boldsymbol{\upsigma}^{\rm bs}\delta_i})\big) \cdot  \mathbf{c}^{\rm med}~~~~~~~
\end{equation*}
\begin{equation}
     T^{\rm obj}_i = \exp\bigg(-\sum_{j=0}^{i-1}\sigma^{\rm obj}_j\delta_j\bigg)~~~~~~~~~~~~~~~~~~~~~~~~
    \label{eqn.def_transmission_revised}
\end{equation}

Based on our derivations we suggest the following architecture (Fig.~\ref{fig:arch}). 
As in~\cite{mildenhall2020nerf}, the object properties are computed by a pair of MLPs, such that the density $\sigma^{\rm obj}$ is a function of the position $(x,y,z)$ only, while color $\mathbf{c}^{\rm obj}$ is determined by the viewing direction $(\theta,\phi)$ as well. The medium parameters $\mathbf{c}^{\rm med}, \boldsymbol{\upsigma}^{\rm bs}, \boldsymbol{\upsigma}^{\rm attn}$ are handled by a separate  MLP with only the direction input, following our decision to constrain them to be constant per viewing direction.






\newcommand{\probP}{\text{I\kern-0.15em P}}
\newcommand{\objnormloss}{\mathcal{L}_{\mathrm{acc}}}
\newcommand{\bsnormloss}{\mathcal{L}_{\mathrm{bs}}}

\subsection{Loss Function}

Let us denote the sequence of samples along a ray by \mbox{$\mathbf{s}=\{s_i\}_{i=0}^N$},  the object `weight' at the $i$th segment  $[s_i,s_{i+1}]$ ,
\begin{equation}
  w^{\rm obj}_i = T^{\rm obj}_i\cdot \big(1-\exp({-\sigma^{\rm obj}_i\delta_i})\big)\;\;,
\end{equation}
the sequence of these weights by $\mathbf{w}=\{w^{\rm obj}_i\}_{i=0}^{N-1}$
and the ground-truth (supervised) pixel color by $C^*$.


Our loss is the following combination: 
\newcommand{\distloss}{\mathcal{L}_{\mathrm{dist}}}
\newcommand{\proploss}{\mathcal{L}_{\mathrm{prop}}}
\newcommand{\reconloss}{\mathcal{L}_{\mathrm{recon}}}
\begin{equation}
\mathcal{L} = \reconloss(\hat{C},C^*) + \proploss(\mathbf{s}, \mathbf{w}) + \lambda\objnormloss(\mathbf{w})\;,
\end{equation}
where $\lambda=0.0001$ was chosen through cross-validation.
Since we work on linear images, we adopt the reconstruction loss of RawNeRF~\cite{mildenhall2022nerf}:
\begin{equation}
\reconloss(\hat{C},C^*) =
\bigg(\frac{\hat{C}-C^*}{\text{sg}(\hat{C})+\epsilon}\bigg)^2
\end{equation}
where sg($\cdot$) stands for stop-gradient and $\epsilon=10^{-3}$.
The `proposal' loss $\proploss(\mathbf{s}, \mathbf{w})$ is an inherent part of Mip-NeRF-360~\cite{barron2022mipnerf360}.
It penalizes for the discrepancy between the distributions of object weights
%
at `original' and `proposed' samplings, where only the latter is used for rendering~\cite{barron2022mipnerf360}.


%

To enforce binary separation between points in space that contain objects and those that contain solely a medium - we add a prior on the transmittance $T^{\rm obj}_i$ of each point on the ray to be either $0$ or $1$, not allowing semi-transparent objects. This is modelled as a mixture of two Laplacian distributions with modes at $0$ and $1$ (following~\cite{rebain2022lolnerf}):
\begin{equation}
\probP(x) \propto e^{-\frac{|x|}{0.1}} + e^{-\frac{|1-x|}{0.1}}
\end{equation}
and use the negative log likelihood loss: 
\begin{equation}
    \objnormloss(\mathbf{w}) = -\log\probP(T^{\rm obj}_i) \;\;.
    \label{eqn:hard2}
\end{equation}




\subsection{Implementation and Optimization}\label{sec:Optimization}
Our implementation is based on the code released in Mip-NeRF-360~\cite{barron2022mipnerf360}, choosing the best performing baseline on our scenes which was  the forward-looking configuration with normalized device coordinates (NDC).
For the mediumMLP , we use 6 linear layers with 256 features and a softplus activation, followed by 3 branches of dense layers and a sigmoid activation for predicting $\boldsymbol{c}^{\rm med}$ and softplus activations for $\boldsymbol{\upsigma}^{\rm attn}$ and $\boldsymbol{\upsigma}^{\rm bs}$. 

In the rendering scheme,~\cite{barron2022mipnerf360} initialized the farthest $\delta_i$ with infinity, enabling the network to predict a background color for rays that do not intersect with any object. We disabled this addition as it prevents our method from explaining the medium that contributes to the rendered color along the ray.
We keep the learning rate and optimization parameters the same as in~\cite{barron2022mipnerf360}. The network is trained for 250,000 iterations with a batch size of $16384$ rays, taking around 10 hours on an Nvidia A100 GPU.  The loss function and metrics are calculated on the  output before any post-processing.














\section{Experiments and Results}
\label{sec:results}

\subsection{Experiments}


\noindent\textbf{Real world scenes.} We acquired multi-image underwater scenes by diving in three different seas: the Red Sea (Eilat, Israel), the Caribbean Sea (Cura\c{c}ao) and the Pacific Sea (Panama) with a total of 20, 20 and 18 images respectively, from which three are set aside for validation in each set. This data encapsulates a diverse set of water conditions and imaging conditions. The images were acquired as RAW images with a Nikon D850 SLR camera in a Nauticam underwater housing with a dome port to avoid refractions that jeopardize the pinhole model~\cite{treibitz2011flat}, and downsampled to average size $900\times 1400$. The input linear images were white-balanced before processing with $0.5\%$ clipping per channel to remove extreme noisy pixels. 
Finally, COLMAP~\cite{schonberger2016structure} is used to extract camera poses.


\noindent\textbf{Simulated Scene.} We constructed a simulation using the \emph{Fern} scene of the LLFF dataset~\cite{mildenhall2019llff}. We ran MIP-NeRF-360~\cite{barron2022mipnerf360} and used the predicted depth maps to simulate an underwater and a foggy scene. Water was added according to~\eqref{eq.scattering} with parameter values: $\beta^D = [1.3, 1.2, 0.9]$, $\beta^B = [0.95, 0.85, 0.7]$, $B^\infty = [0.07, 0.2, 0.39]$.
Fog was simulated based on~\cite{sakaridis2019guided} with $\beta = 1.2$. 

\begin{figure}[t]
  \centering
  \includegraphics[width=1\linewidth]{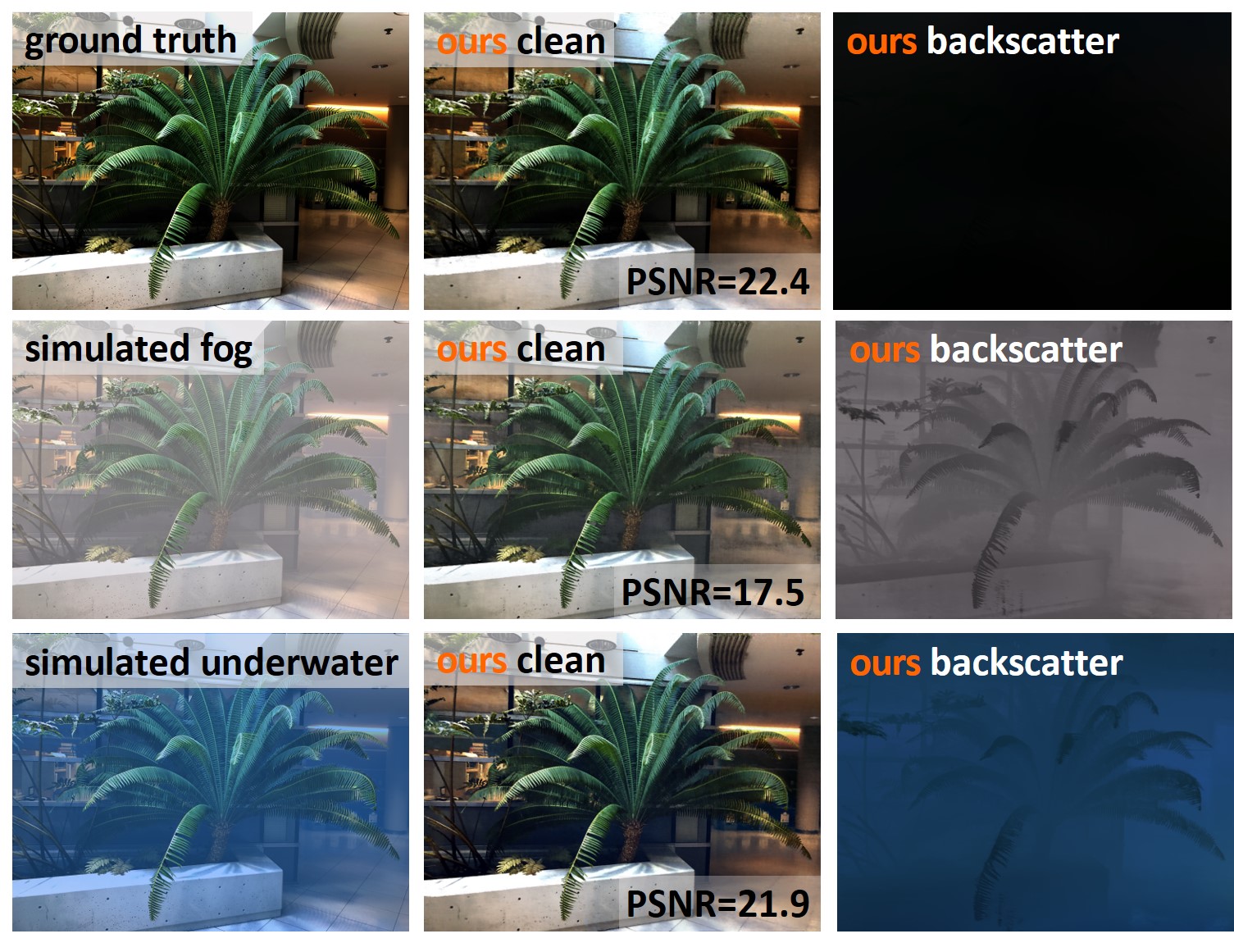} \vspace{-18pt}
   \caption{\textbf{Synthetic experiment on \emph{Fern} scene of the LLFF dataset~\cite{mildenhall2019llff}.} \textbf{\textit{Top}}: Our method can handle `in-air' scenes, with a uniform zeros backscatter image and a reconstruction PSNR that is close to that of the baseline~\cite{barron2022mipnerf360} which was 23.73. \textbf{\textit{Middle}} and \textbf{\textit{bottom}}: Our method's separation between clean and backscatter components, for simulations of fog and underwater effects. The reconstruction quality degrades rather gracefully.}\vspace{-0.25cm}
   \label{fig:sim}
\end{figure}



\begin{figure*}[t]
  \centering
  \includegraphics[width=\linewidth]{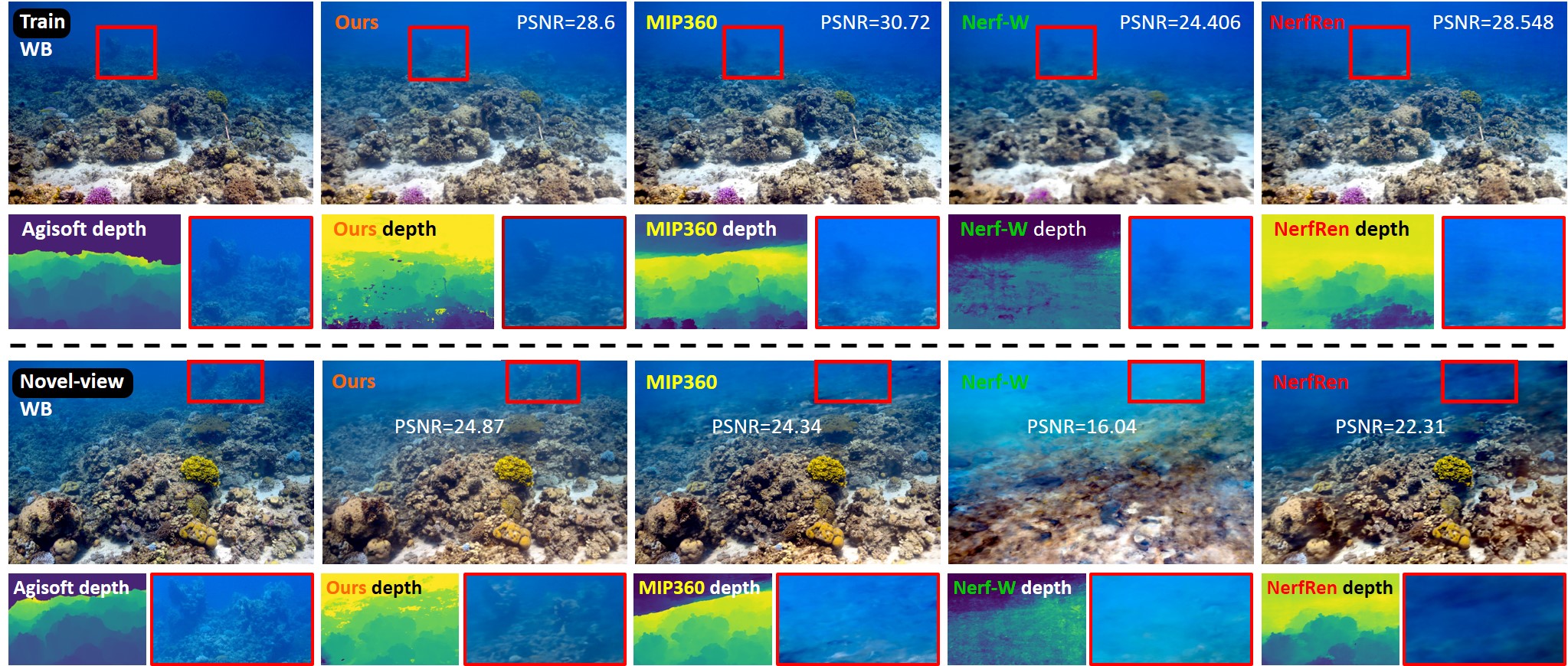} \vspace{-17pt}
   \caption{\textbf{Scene rendering in the medium, on `Red Sea'}. Left to right: white-balanced input image, our result, MIP360~\cite{mildenhall2022nerf}, Nerf-W~\cite{martin2021nerf}, NeRFReN~\cite{guo2022nerfren}. \textbf{[Top 2 rows]}~Train image and zoom-ins.  Even in the task of overfitting to the training images, there are noticeable differences between the methods, and as can be seen in the zoom-ins (red squares)  we are able to reconstruct fine details in further areas (albeit a lower PSNR). \textbf{[Bottom 2 rows]}~Novel view and zoom-ins. Our method achieves the highest PSNR (see tables~\ref{tab:uw_psnr},~\ref{tab:uw_psnr2}), and provides much better details in further areas (see red square zoom-ins). Our depth reconstructions provide more detail in further areas.
   }\vspace{-0.2cm}
   \label{fig:uw_nerf_qual}
\end{figure*}

\noindent\textbf{Baseline methods.} The input to all methods is the same set of white-balanced linear images.
In the task of \textbf{rendering scenes in the medium} we compare with the following NeRF methods: MIP-NeRF-360, forward-looking with NDC (MIP360)~\cite{barron2022mipnerf360}, NeRF-W~\cite{martin2021nerf} and NeRFReN~\cite{guo2022nerfren}. 
In the task of \textbf{reconstructing clean medium-free images} we compare with NeRF-W, where the `transient' component can be viewed as the clean image and the `static' image as backscatter, and to NeRFReN with the `reflected' and `transmitted' components as clean and backscatter images. 
We additionally compare to single-image scene reconstruction methods: plain white balance,  SeaThru~\cite{akkaynak2019sea}, the leading ``classical'' methods of Bekerman et al~\cite{bekerman2020unveiling}, and the recent deep-learning based ~\cite{zhang2022underwater} (as well as \cite{han2022underwater,zhang2022underwater} whose results are provided in the project website).  
SeaThru requires depth maps, that we generate using SFM (with Agisoft Metashape) as originally suggested. These depth-maps are used to compare with those of the different methods. 

\noindent\textbf{Photo-finishing:} As the input and output images are linear, we apply photofinishing on all linear reconstructed scenes to enhance scene contrast and appearance, using the digital camera pipeline from~\cite{karaimer2016software}. This is done to improve visualization for easier qualitative comparisons, while 
PSNR is calculated on the original non-photofinished linear images. 


\subsection{Results}



\noindent We start with a \textbf{sanity check}  on a clean image. Our method correctly estimates  zero  backscatter, and does not force itself to estimate a medium (Fig.~\ref{fig:sim}~Top). On our \textbf{simulated} fog and underwater scenes (Fig.~\ref{fig:sim}~middle and bottom), our method separates the scenes' components very well, with only slight reduction in PSNR in the underwater scene.

\noindent\textbf{Image rendering and novel-view synthesis (in medium).} Qualitative results are summarized in Figs.~\ref{fig:fig1},~\ref{fig:uw_nerf_qual}. Table~\ref{tab:uw_psnr} summarizes the average PSNR on the validation set of the Red Sea scene, in which our method achieves the highest PSNR. Note that both NeRF-W~\cite{martin2021nerf} and NeRFReN~\cite{guo2022nerfren} are based on the NeRF~\cite{mildenhall2020nerf} model while our code is based on Mip-NeRF-360~\cite{barron2022mipnerf360} that in general improves reconstruction details in further areas. So our real comparison is with~\cite{barron2022mipnerf360}, which we improve by  $\sim 0.8dB$ on average over the validation set. 
Table~\ref{tab:uw_psnr2} compares PSNR, SSIM and LPIPS with those of~\cite{barron2022mipnerf360} on validation sets of all real-world and simulation scenes. Our method is better in the majority of cases, and is \emph{especially good on the further areas (red squares)}. 

On the train set ~\cite{barron2022mipnerf360} achieves a reasonable rendering of the scene in terms of PSNR by \textbf{wrongly modeling the water as a nearby blue object}, as can be seen in the depth map, indicating a close object in the top area.
Even then, in our results  further objects are reconstructed with more detail. 
The NeRFReN~\cite{guo2022nerfren} depth map flattens at the mid-range, while ours is more informative further on. SFM does not estimate depth at these areas because of lack of features. 

\begin{table}[t]
\setlength{\tabcolsep}{6pt}
\begin{center}
\small
\begin{tabular}{|c|c|c|c|c|c|c|}
\hline
 \textbf{Ours} & I  & II & III & \cite{barron2022mipnerf360} & \cite{martin2021nerf}  & \cite{guo2022nerfren}  \\\hline
\textbf{21.83} & 21.76  & 21.43 & 21.72 & 21.05 & 16.52  & 21.05  \\\hline
\end{tabular}
\vspace{-5pt}
   \caption{\textbf{Average PSNRs of in-medium rendering on Red Sea  set}: ours, 3 ablation variations (description in the text: I-- 1 parameter, II-- 3 parameters, III-- Eqs.~\ref{eq.color_interval_UW_obj}, \ref{eq.color_interval_UW_med}) and other methods. }
   \label{tab:uw_psnr}
\end{center}\vspace{-0.4cm}
\end{table}

\begin{table}[t]
\setlength{\tabcolsep}{5pt}
\begin{center}
\small
\begin{tabular}{|l|c|c|}
\hline
 \textbf{Scene} & \textbf{MIP360 \cite{barron2022mipnerf360}}    & \textbf{Ours} \\\hline
Red Sea  & 21.05 / 0.75 / 0.29  & \textbf{21.83} / \textbf{0.77} / \textbf{0.25}   \\\hline
Red Sea red square & 29.66 / 0.84 / 0.43   & \textbf{33.80} / \textbf{0.90} / \textbf{0.23}   \\\hline
Cura\c cao     & 26.54 / 0.81 / 0.33  & \textbf{30.48} / \textbf{0.87} / \textbf{0.20} \\\hline
Cura\c cao red square & 27.04 / 0.84 / 0.45   & \textbf{33.20} / \textbf{0.88} / \textbf{0.08} \\\hline
Panama      & 27.43 / 0.82 / 0.23  & \textbf{27.89} / \textbf{0.83} / \textbf{0.22}  \\\hline
Fern fog   &  30.23 / \textbf{0.88} / \textbf{0.15}  & \textbf{30.75} / 0.87 / \textbf{0.16} \\\hline
Fern underwater   &  29.62 / \textbf{0.87} / 0.26  & \textbf{29.76} / 0.86 / \textbf{0.15} \\\hline
\end{tabular}\vspace{-5pt}
\caption{\textbf{Comparison to baseline}: PSNR$\uparrow$ / SSIM$\uparrow$ / LPIPS$\downarrow$ .}
\label{tab:uw_psnr2}
\end{center}\vspace{-0.8cm}
\end{table}

\begin{figure*}[t]
  \centering
 \includegraphics[width=1\linewidth]{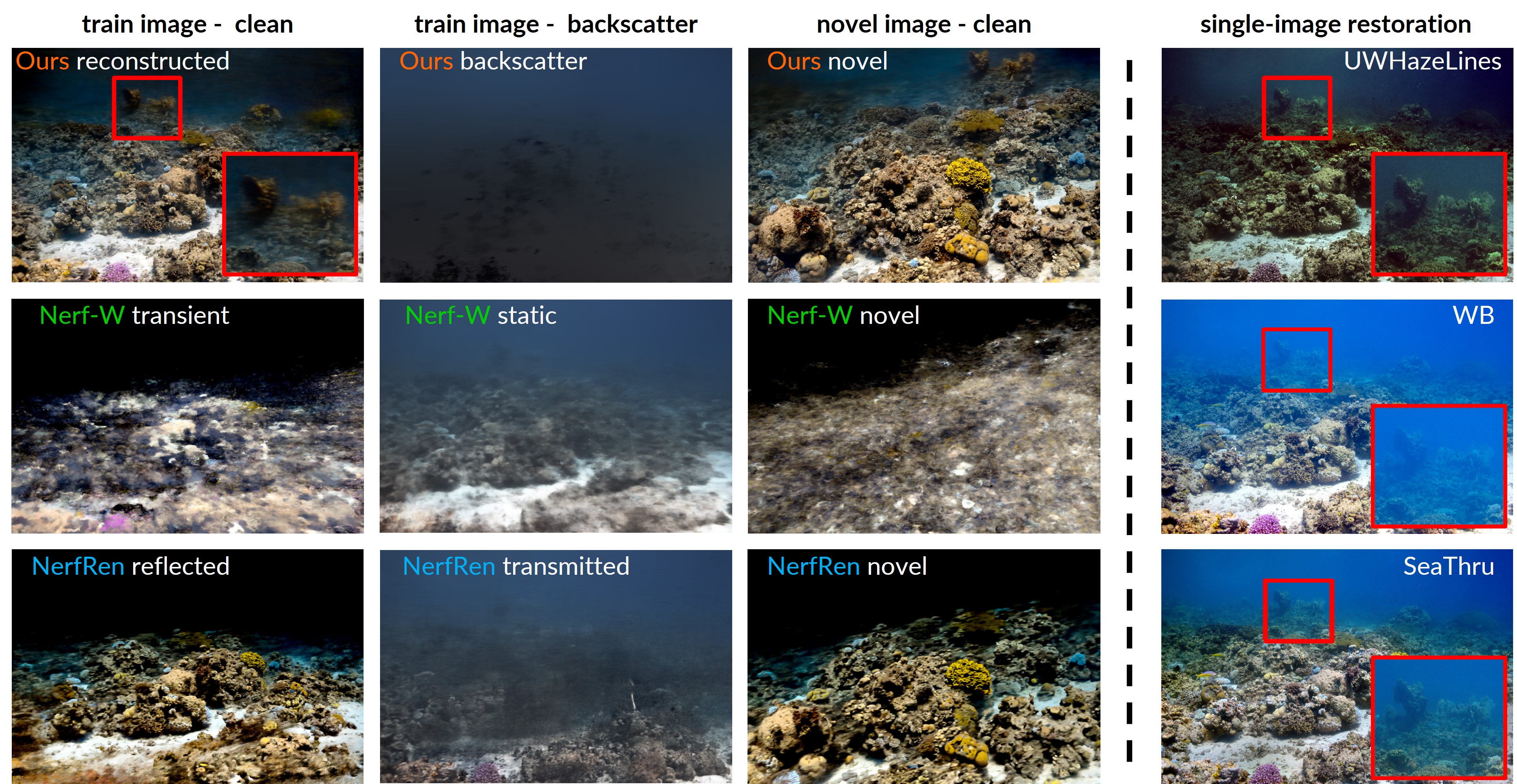} \vspace{-17pt}
   \caption{\textbf{Scene reconstruction on `Red Sea'}. \textit{\textbf{Columns 1-3}}: Comparing our method (top row) and 2 other NeRFs - Nerf-W~\cite{martin2021nerf} (middle row) and NeRFReN~\cite{guo2022nerfren} (bottom row). Even though  not developed for scattering media, they manage to separate a training image into two components (columns 1, 2), resembling the clean scene and the backscatter. However, their backscatter components contains details of the objects, while ours correctly displays only the signal stemming from the medium. Similarly (column 3), their 'clean' rendering of a novel view is inferior to ours, especially in the far distances. \textit{\textbf{Column 4}}: Among the single-image methods, SeaThru~\cite{akkaynak2019sea} is the best performing (but  requires the depth map as input). Nevertheless, color artifacts are visible in the far parts due to inaccurate depth information. 
   }
   \label{fig:ours_clean}\vspace{-10pt}
\end{figure*}


\noindent\textbf{Scene restoration (Fig.~\ref{fig:ours_clean}).} In comparison with NeRF methods for scene separation it is clear that our separation handles the medium the best. In the result of NeRF-W~\cite{martin2021nerf} the scene is somewhat separated but blurry, while in that of NeRFReN~\cite{guo2022nerfren} the closer part of the scene is relatively well reconstructed but the further areas are missing. In both, the scene's objects are visible in the backscatter image, which is clearly wrong and reduces relevant signal from the object image. In our results, due to the adequate modeling, the backscatter is not polluted with objects. Compared to scene restoration methods, SeaThru~\cite{akkaynak2019sea} is the best performing one but is limited to areas where the depth map is available from SFM. The single image methods are not on par, especially from the mid-ranges and on. In \textbf{novel-view synthesis of clean scenes} we are able to  predict further the entire scene.



\vspace{5pt}
\noindent\textbf{Ablation Study.} (i) \emph{Number of medium parameters:} following the discussion in Sec.~\ref{sec:uw_back}, the elaborate version of Eq.~\eqref{eq.scattering} uses 6 parameters for the medium coefficients, while simplified versions use 1 (marked I in Table~\ref{tab:uw_psnr}) or 3 (marked II). This is ablated in Table~\ref{tab:uw_psnr}, in terms of PSNR, where the elaborate model with 6 parameters achieves a higher score. 
(ii) \emph{Rendering equations.} We compare our basic rendering model  (Eqs.~\ref{eq.color_interval_UW_obj}, \ref{eq.color_interval_UW_med}), marked III in Table~\ref{tab:uw_psnr} to the final one marked `ours' (Eq.~\ref{eqn.def_transmission_revised}). This option achieves lower PSNR.

\section{Discussion}\vspace{-0.1cm}
\label{sec:discussion}

We provide an important extension of  NeRFs  that enables rendering scenes acquired in scattering media such as haze, fog and underwater.  So far, NeRF provided a framework for volumetric rendering, but without considering the nature of the medium, resulting in a binary `occupancy' volume. 
Our formulation enables  opaque objects to exist in a semi-transparent medium that is both scattering and absorbing in a wavelength-dependent manner. We demonstrate it on challenging real-world scenes with a complex 3D structure. Our scenes are forward-facing and contain areas with no objects at all, which our method is able to explain. 

Water effects in further regions are very strong, hampering single image methods and feature matching in multiview methods. Our  method incorporates information from all images at once, learning the scene in a holistic way. This enables better scene reconstruction (in medium and clean), and estimating the depth and the medium's parameters. 

Our method has several limitations. While it is based on the current state-of-the-art image formation model, that model does not account for multiple scattering or for artificial illumination. As common in NeRFs, it requires  camera poses that are extracted beforehand, which can be challenging in  bad visibility.  Lastly, the medium's parameters are better learned in sets where there is enough variation in the scene range between the viewpoints.  
The formulation takes its strength from the modeling of the medium. Thus it struggles in scenes that do not adhere to the model's assumptions, e.g., underwater scenes with significant flickering. In the future we plan to add components that will explain transient effects such as flickering, and continue to explore estimation of more diverse scenes and medium parameters.

\noindent {\small \textbf{Acknowledgements.} The research was funded by Israel Science Foundation grant $\#680/18$, Israeli Ministry of Science and Technology grants $\#1001577600$ \& $\#1001593851$,  EU Horizon 2020 research and innovation programme GA $101094924$ (ANERIS), the  Leona M. and Harry B. Helmsley Charitable Trust, the Maurice Hatter Foundation, and the Schmidt Marine Technology Partners. We thank Matan Yuval for substantial data contribution, Opher Bar-Nathan and Yuval Goldfracht for help with experiments.
}


{\small
\bibliographystyle{ieee_fullname}
\bibliography{egbib}
}


\end{document}